\documentclass[10pt,twocolumn,letterpaper]{article}

\usepackage[pagenumbers]{wacv} 

\usepackage{graphicx}
\usepackage{amsmath}
\usepackage{amssymb}

\usepackage{caption}
\usepackage{subcaption}
\usepackage{enumerate}
\usepackage{wrapfig}
\usepackage{lipsum}
\usepackage{breqn}
\usepackage{chngcntr}
\usepackage{booktabs}
\usepackage{etoc}
\usepackage{color}
\usepackage{nicefrac}
\usepackage{verbatim}
\usepackage{multicol}
\usepackage{multirow}
\usepackage{pifont}
\newcommand{\cmark}{\ding{51}}%
\usepackage{graphicx}
\usepackage{caption}
\usepackage{shorttoc}
\usepackage{enumitem}
\usepackage{appendix}
\usepackage{bbm}
\usepackage{fancyhdr}

\newcommand{\interval}[1]{\scriptsize{$\pm$ #1}}
\newcommand{\smallinterval}[1]{\tiny{$\pm$ #1}}

\def\modelName{{DiaMond}}

\fancypagestyle{firstpagestyle}{
    \fancyhf{} 
    \fancyhead[R]{Accepted by WACV 2025}
    
}

\usepackage[pagebackref,breaklinks,colorlinks]{hyperref}

\usepackage[capitalize]{cleveref}
\crefname{section}{Sec.}{Secs.}
\Crefname{section}{Section}{Sections}
\Crefname{table}{Table}{Tables}
\crefname{table}{Tab.}{Tabs.}

\def\wacvPaperID{75} 
\def\confName{WACV}
\def\confYear{2025}

\begin{document}

\title{DiaMond: Dementia Diagnosis with Multi-Modal Vision Transformers \\ Using MRI and PET}


\author{\textbf{Yitong Li$^{1,2}$} \quad \textbf{Morteza Ghahremani$^{1,2}$} \quad \textbf{Youssef Wally$^{1}$} \quad \textbf{Christian Wachinger$^{1,2}$} \vspace{0.3em} \\
{\normalsize $^1$Lab for Artificial Intelligence in Medical Imaging, Technical University of Munich (TUM), Germany} \quad  \\
{\normalsize $^2$Munich Center for Machine Learning (MCML), Germany} \\
\tt \normalsize yi_tong.li@tum.de
}

\maketitle

\begin{abstract}

\thispagestyle{firstpagestyle}

Diagnosing dementia, particularly for Alzheimer's Disease (AD) and frontotemporal dementia (FTD), is complex due to overlapping symptoms. 
While magnetic resonance imaging (MRI) and positron emission tomography (PET) data are critical for the diagnosis, 
integrating these modalities in deep learning faces challenges, often resulting in suboptimal performance compared to using single modalities.
Moreover, the potential of multi-modal approaches in differential diagnosis, which holds significant clinical importance, remains largely unexplored.
We propose a novel framework, \modelName, to address these issues with vision Transformers to effectively integrate MRI and PET. \modelName\ is equipped with self-attention and a novel bi-attention mechanism that synergistically combine MRI and PET, alongside a multi-modal normalization to reduce redundant dependency, thereby boosting the performance. 
\modelName\ significantly outperforms existing multi-modal methods across various datasets, achieving a balanced accuracy of 92.4\% in AD diagnosis, 65.2\% for AD-MCI-CN classification, and 76.5\% in differential diagnosis of AD and FTD. 
We also validated the robustness of \modelName\ in a comprehensive ablation study.
The code is available at \url{https://github.com/ai-med/DiaMond}.


\end{abstract}    
\section{Introduction}

\begin{figure}[t]
\vspace{-1.5em}
    \centering
    \includegraphics[width=\linewidth]{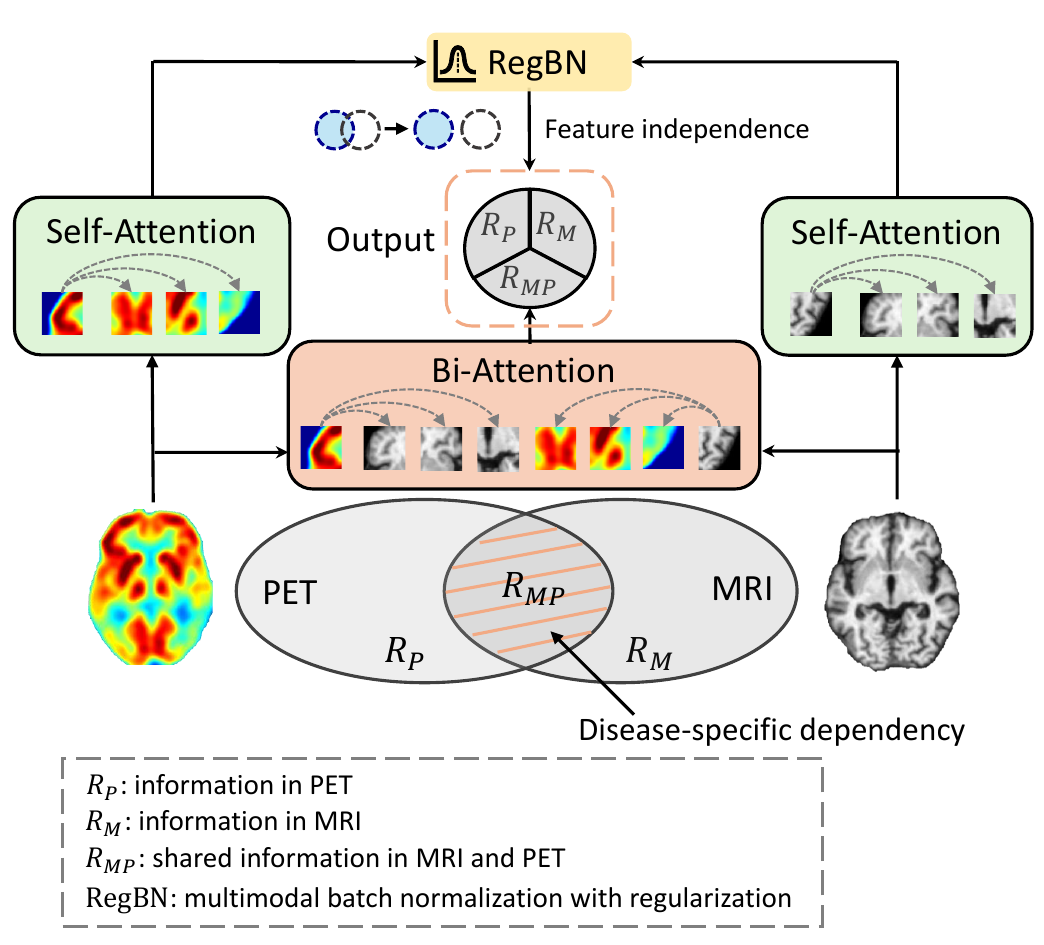}
    \caption{MRI and PET are two modalities with disease-specific dependency. We introduce a novel framework including self-attention mechanism with multi-modal normalization to capture distinct features from single modalities, and a novel bi-attention mechanism to exclusively extract their similarities.} 
    \label{fig:intro}
    \vspace{-1.5em}
\end{figure}

Dementia presents a growing concern for individuals and society, with Alzheimer’s Disease (AD) constituting 60-80\% of the cases and frontotemporal dementia (FTD) ranking as the second most common type in the younger-elderly population under 65 years old~\cite{Young2018FrontotemporalDL}. 
Accurately diagnosing AD and distinguishing it from other dementia types is crucial for patient management, therapy, and prognosis, but challenging as symptoms overlap. 
To address this challenge, a variety of diagnostic tools are employed, including magnetic resonance imaging (MRI), positron emission tomography (PET), and cognitive tests~\cite{addiagnosis}.
Structural MRI provides details on neuroanatomy for identifying regional atrophy, while fluorodeoxyglucose (FDG) PET tracks the distribution of glucose metabolism in the brain. 
Recent research into machine learning methods has revealed that using both MRI and PET scans improves the accuracy of diagnosing 
AD and distinguishing it from other types of dementia, compared to single-modality imaging~\cite{dukart2011combined}.
In deep learning (DL), CNNs have been widely employed to combine these two modalities, but recent results pointed out that the combination may not surpass the efficacy of using PET alone  \cite{petallyouneed}. 
The emergence of Transformers and attention mechanisms gives rise to a surge of new techniques in multi-modal AD diagnosis~\cite{Zhang2023transformer,GAO2023incomplete,tang2024multimodal}, yet they commonly integrate CNNs for initial feature extraction with vision Transformers (ViTs) for feature fusion, which may not fully exploit the multi-modal potential of ViTs. 
Moreover, although the main clinical objective for neuroimaging of dementia patients is differential diagnosis, current studies have been limited to consider one single type of dementia, namely AD~\cite{cnnlstm2019,earlyfusion,huang2019diagnosis}.

We address this gap by introducing \modelName, a novel framework that leverages pure ViTs with multiple attention mechanisms for effective multi-modal classification, which exhibits high efficacy in both AD prediction and differential diagnosis of dementia. 
\modelName\ consists of independent branches incorporating \emph{self-attention} and \emph{bi-attention} mechanisms, where the former extracts unique features from PET and MRI modalities independently while the latter exclusively captures correlations between modalities in their overlapping information, as illustrated in Fig.~\ref{fig:intro}. 
Since the bi-attention block explicitly focuses on the potential similarity between modalities, we remove the redundant partial dependencies between modalities from the self-attention computation using the multi-modal normalization technique RegBN~\cite{ghahremani2024regbn}, enabling these branches to efficiently explore distinct features within each modality. 
Evaluated on three distinct medical datasets, \modelName\ shows significant improvement not only in the classification of  AD progression but also in the differential diagnosis between AD and FTD.
In summary, our contributions are:
\begin{itemize}[label=\textbullet]
    \item An efficient multi-modal classification framework using pure ViTs for feature extraction and interaction.
    \item A novel bi-attention mechanism to exclusively explore the underlying similarities between PET and MRI data in high-dimensional feature space. 
    \item Integrating the multi-modal normalization technique RegBN to reduce redundant dependencies between multiple modalities for enhanced diagnosis accuracy.
    \item Robustness of DiaMond evaluated on three distinct datasets for AD prediction and differential diagnosis of AD and FTD. DiaMond is the first method to leverage deep neural networks to enhance differential diagnosis of dementia with MRI and PET.
\end{itemize}






\section{Related Work}

\noindent\textbf{Multi-Modal Learning for AD Diagnosis.} Recent studies have advanced AD diagnosis by combining structural MRI and FDG-PET through various DL approaches.
Different fusion strategies have been explored, including early~\cite{earlyfusion,lin2021bidirectional,huang2019diagnosis, wen2024high}, middle~\cite{petallyouneed,zhou2019effective}, and late fusion~\cite{petallyouneed,huang2019diagnosis,cnnlstm2019,zhou2019effective}. 
Lu \etal~\cite{lu2018multimodal} use a multiscale deep neural network to fuse the extracted patch-wise 1D features from MRI and PET.
Liu \etal~\cite{liu2018multi} propose a cascaded framework including multiple deep 3D-CNNs to learn from local image patches and an upper 2D-CNN to ensemble the high-level features.
Feng \etal~\cite{cnnlstm2019} combine a 3D CNN and LSTM with a late fusion of MRI and FDG-PET for AD diagnosis.
Huang \etal~\cite{huang2019diagnosis} propose an early and a late fusion approach for the two modalities based on a 3D-VGG.
Lin \etal~\cite{lin2021bidirectional} first propose a 3D reversible GAN for imputing missing data and then use a 3D CNN to perform AD diagnosis with channel-wise early fused MRI and PET data.
Wen \etal~\cite{wen2024high} introduce an adaptive linear fusion method for MRI-PET fusion based on 2D CNNs.
Song \etal~\cite{earlyfusion} propose an early fusion approach by overlaying gray matter (GM) tissues from MRI with the FDG-PET scans and feeding them into a 3D CNN for classification.
However, after investigating several multi-modal methods with 3D CNN across image-level early, middle, and late fusion of MRI and PET, Narazani \etal~\cite{petallyouneed} find that the diagnostic performance of these existing multi-modal fusion techniques may not yet outperform that of using PET alone for AD diagnosis. 

As Transformers and attention mechanisms have shown promising results in various medical imaging tasks~\cite{he2023h2former,chen20233d,ghahremani2024h}, recent applications have started to integrate them for multi-modal feature fusion with CNNs as encoders for feature extraction~\cite{ZHANG2023attention,KADRI2023coat,LI2023dfenet,GAO2023incomplete,Zhang2023transformer,mmtfn2024,tang2024multimodal}.
Li \etal~\cite{LI2023dfenet} combine a CNN and a Transformer module for multi-modal medical image fusion.
Zhang \etal~\cite{ZHANG2023attention} propose an end-to-end 3D ResNet framework, which integrates multi-level features obtained by attention mechanisms to fuse the features from MRI and PET.
Gao \etal~\cite{GAO2023incomplete} introduce a multi-modal Transformer (Mul-T) using DenseNet and spatial attention for global and local feature extraction, followed by cross-modal Transformers for T1-, T2-MRI, and PET fusion.
Zhang \etal~\cite{Zhang2023transformer} use MRI and PET for dementia diagnosis by first employing adversarial training with CNN encoders for feature extraction, then applying Transformers through the cross-attention mechanism for feature fusion and finally classification with a fully-connection layer.
Miao \etal~\cite{mmtfn2024} propose a multi-modal multi-scale transformer fusion network (MMTFN), combining CNN-based residual blocks and Transformers to jointly learn from multi-modal data for diagnosing AD.
Tang \etal~\cite{tang2024multimodal} first employ a 3D CNN to extract deep feature representations of structural MRI and PET images, then utilize an improved Transformer to progressively learn the global correlation information among features. 
Note that these recent approaches typically combine CNNs for initial feature extraction with ViTs for feature fusion, which may not fully leverage the capabilities of ViTs for multi-modal learning.
\vspace{0.5em}

\begin{figure*}[t]
\vspace{-1em}
    \centering
    \includegraphics[width=\linewidth]{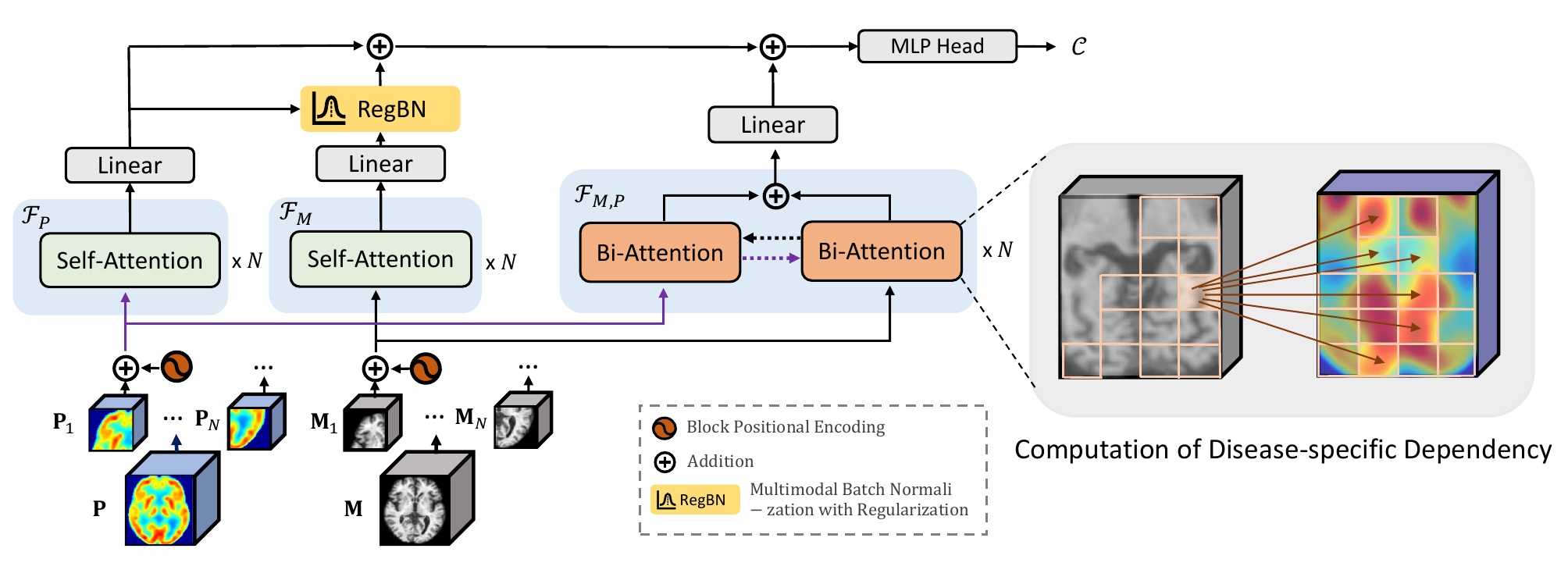}
    \caption{\modelName\ encodes PET and MRI individually with blocks $\mathcal{F}_{P}$ and $\mathcal{F}_{M}$, then applies RegBN to ensure their independence. It leverages a novel bi-attention mechanism on the multi-modal features in block $\mathcal{F}_{M,P}$ to focus on correlations among them. Finally, the latent features from all three blocks are aggregated into an MLP head to obtain classification labels $\mathcal{C}$.
    }
    \label{fig:framework}
    \vspace{-1em}
\end{figure*}

\noindent\textbf{Differential Diagnosis of Dementia.} Despite the extensive research focused on AD diagnosis, there has been limited exploration in multi-modal learning for differential diagnosis of dementia, which holds significant importance in clinical practices~\cite{chouliaras2023use}. 
Current research in differential diagnosis primarily adopts machine learning algorithms~\cite{dukart2011combined,gjerum2020evaluating,cagnin2024data} or focuses on single modality~\cite{libarlow,nguyen2022interpretable}, yet the potential of employing multi-modal deep neural networks for this purpose remains unexplored.

\section{Proposed Method}
\label{sec:Methods}

\cref{fig:framework} presents the main steps of \textit{DiaMond}, a multi-modal ViT-based framework incorporating multiple attention mechanisms for effective dementia diagnosis. We begin by outlining the foundational concepts underlying our approach and subsequently delve into the specifics of each component.

\subsection{Preliminaries}\label{sec:prelim}
Let's denote the 3D MRI and PET images as $\mathbf{M}\in{\mathbb{R}^{H\times W\times D}}$ and $\mathbf{P}\in{\mathbb{R}^{H\times W\times D}}$ respectively, with height $H$, width $W$, and depth $D$. The objective is to classify the given multi-modal data into a set of $L$ labels $\mathcal{C}=\{{c}_1,\cdots,{c}_L\}$. In this study, $\mathcal{C}$ includes CN (Cognitively Normal), MCI (Mild Cognitive Impairment), AD, and FTD labels. 
MRI and PET data exhibit inherent dependencies that can be introduced during data collection and revealed in high-dimensional feature space. Hence, we divide the data space into three non-overlapping regions denoted by $\{R_M, R_P, R_{MP}\}$, as highlighted in ~\cref{fig:intro}.
Our proposed framework \modelName\ consists of three branches based on pure ViTs to process each of the data space: 
\begin{enumerate}
    \item $\mathcal{F}_{M}: \mathbf{M}\in{\mathbb{R}^{H\times W\times D}}\rightarrow z_{M}\in{\mathbb{R}^{f}}$ maps input MRI from data space $R_M$ (Fig.~\ref{fig:intro}) into the latent encoding $z_{M}$ with length $f$;
    \item $\mathcal{F}_{P}: \mathbf{P}\in{\mathbb{R}^{H\times W\times D}}\rightarrow z_{P}\in{\mathbb{R}^{f}}$ maps PET from data space $R_P$ into the latent encoding $z_{P}$ of length $f$; 
    \item $\mathcal{F}_{{M,P}}: \{\mathbf{M},\mathbf{P}\}\rightarrow z_{M,P}\in{\mathbb{R}^{f}}$ receives both MRI and PET, then captures their shared information in data space $R_{MP}$, and finally maps those to the latent encoding $z_{M,P}$ of length $f$. 
\end{enumerate}
We extract regions $\{R_M, R_P\}$ using self-attention mechanisms, together with a recently-developed normalization technique RegBN~\cite{ghahremani2024regbn} to ensure feature independence. A novel bi-attention mechanism is introduced to explore the similarities between the two modalities in region $R_{MP}$.

\subsection{Self-Attention for Single Modality}
$\mathcal{F}_{{M}}$ and $\mathcal{F}_{{P}}$ operate over single modalities separately with self-attention mechanisms. 
Let $\mathbf{M}$ and $\mathbf{P}$ be partitioned into $h\times w\times d$ voxel patches $\mathbf{X}^{m}\in \mathbb{R}^{N\times h\times w\times d\times f_{e}}, m\in\{\mathbf{M},\mathbf{P}\}$, where $f_{e}$ denotes the length of the feature embedding, and $N$ the number of input patches. 
The input patch $\mathbf{X}^{m}$ is projected into the query $\mathbf{W}^{m}_{Q}\in \mathbb{R}^{f_{e}\times f_{e}}$, key $\mathbf{W}^{m}_{K}\in \mathbb{R}^{f_{e}\times f_{e}}$, and value $\mathbf{W}^{m}_{V}\in \mathbb{R}^{f_{e}\times f_{e}}$ matrices, giving:
\begin{multline}\label{eq::qkv}
    \mathbf{Q}^{m}=\mathbf{X}^{m}\mathbf{W}^{m}_{Q}, \:\:
    \mathbf{K}^{m}=\mathbf{X}^{m}\mathbf{W}^{m}_{K}, \:\:
    \mathbf{V}^{m}=\mathbf{X}^{m}\mathbf{W}^{m}_{V}, \\
 \quad m \in \{\mathbf{M},\mathbf{P}\}.
\end{multline}
Let $\top$ denote transpose, the self-attention $\mathbf{A}_{\text{self}}$ for a single modality $m$ is defined as 
\begin{equation}\label{eq:sself}
\begin{split}
    \mathbf{A}_{\text{self}}\left(\mathbf{X}^m\right) 
    &= \texttt{attention} \left(\mathbf{Q}^m,\mathbf{K}^m,\mathbf{V}^m\right) \\
    &= \texttt{softmax}\left(\frac{\mathbf{Q}^m\mathbf{K}^{m,\top}}{\sqrt{f_e}}\right)\mathbf{V}^m.
\end{split}
\end{equation}

\noindent Each self-attention branch aims to independently extract unique features from one input modality. To ensure that each branch efficiently identifies distinct modality-dependent features, a normalization technique RegBN~\cite{ghahremani2024regbn} is later applied to the latent space, aiming to reduce redundant partial dependency between the self-attention branches.

\subsection{Bi-Attention for Multiple Modalities}
\begin{figure}[t]
 \vspace{-1em}
    \centering
    \includegraphics[width=1\linewidth]{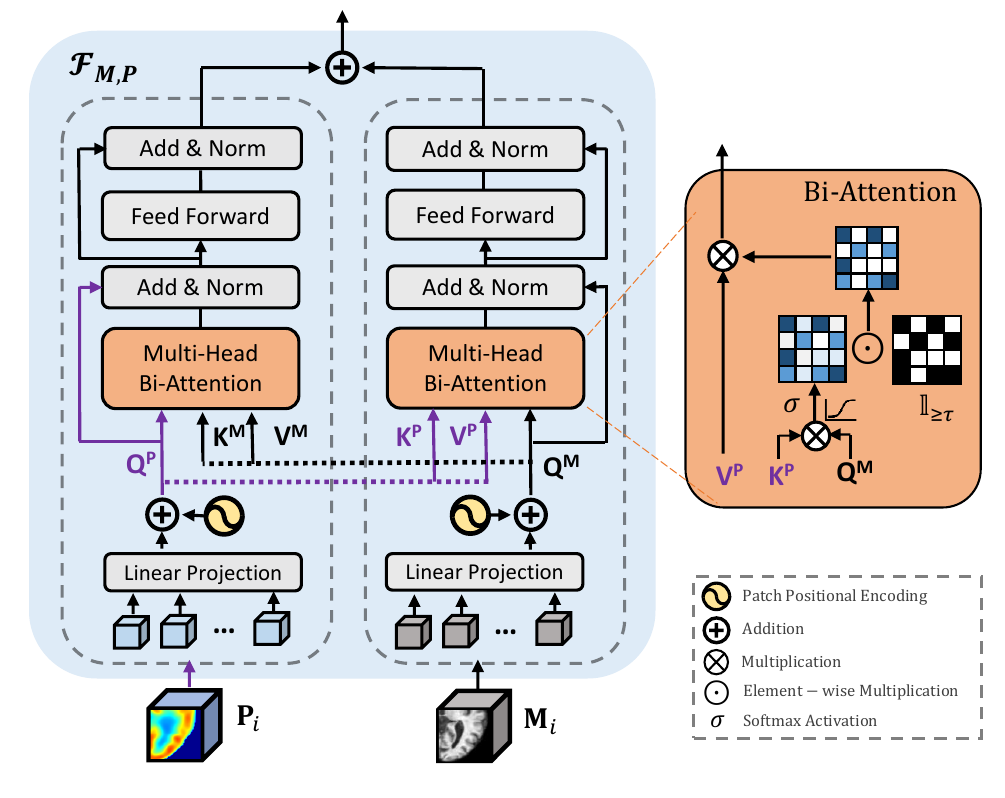}
    \vspace{-1em}
    \caption{Bi-attention block in branch $\mathcal{F}_{{M,P}}$ calculates the interwaved attention between input modalities, with a constant threshold $\tau$ to filter out very small values in the correlation matrices.}
    \label{fig:bi-attention}
    \vspace{-1.1em}
\end{figure}

A novel bi-attention mechanism $\mathbf{A}_{\text{bi}}$ is introduced in $\mathcal{F}_{{M,P}}$ to compute the interweaved attention between two modalities $\{m,n\}$, uniquely designed to focus on capturing their similarities in the high-dimensional feature space: 
\begin{equation}\label{eq:biattention}
\begin{split}
    \mathbf{A}_{\text{bi}}\left(\mathbf{X}^m,\mathbf{X}^n\right) &= \texttt{attention}_{\tau} \left(\mathbf{Q}^m,\mathbf{K}^n,\mathbf{V}^n\right) \\
    &= \mathbbm{1}_{\left(\mathbf{Z}^{m,n} \geq \tau \right)} \cdot \mathbf{Z}^{m,n} \cdot \mathbf{V}^n,
\end{split}
\end{equation}
\begin{equation}\label{eq:zmn}
    \mathbf{Z}^{m,n} = 
    \texttt{softmax} \left(\frac{\mathbf{Q}^m\mathbf{K}^{n,\top}}{\sqrt{f_e}}\right),
\end{equation}
where $ m \in \{\mathbf{M},\mathbf{P}\}$, $n \in \{\mathbf{P},\mathbf{M}\}$, and $m \neq n$. $\mathbbm{1}_{(\cdot)}$ is the indicator function to threshold the correlation matrix between the features of the two modalities to be above a constant threshold $\tau$.
Note that $\mathbf{A}_{\text{bi}}\left(\mathbf{X}^{\mathbf{M}},\mathbf{X}^{\mathbf{P}}\right) \neq \mathbf{A}_{\text{bi}}\left(\mathbf{X}^{\mathbf{P}},\mathbf{X}^{\mathbf{M}}\right)$. Illustrated in~\cref{fig:bi-attention}, the bi-attention blocks aim to produce features for each modality conditioned on the other, targeting on their potential disease-specific similarities. 
Distinct from the conventional cross-attention mechanism, which captures both similarities and dissimilarities between patches in the correlation matrices, our proposed bi-attention selectively preserves only the pronounced similarities, resulting in a sparse representation. This sparsity is achieved by applying a constant threshold $\tau$ to the correlation matrix between the query and key matrices from two modalities, filtering out negligible values.
This mechanism enables efficient capture of dependencies between modalities for improved diagnostic accuracy and robustness, as validated in Sec.~\ref{sec:ablation_branches} \& \ref{sec:ablation_tau}.

\subsection{RegBN for Dependency Removal}
RegBN is a normalization method devised for dependency and confounding removal from low- and high-level features before fusing those~\cite{ghahremani2024regbn}. 
As discussed earlier, MRI and PET input images are mapped into latent space, represented as $z_{M}\in{\mathbb{R}^{f}}$ for MRI and $z_{P}\in{\mathbb{R}^{f}}$ for PET. 
Dependencies between modalities are present during image acquisition; thus, such information can be transferred to the latent space, as illustrated by $R_{MP}$ in Fig.~\ref{fig:intro}. Since the proposed bi-attention block focuses explicitly on the underlying similarities between input modalities via the self-attention modules, it is essential to eliminate redundant shared information between them. Otherwise, the neural network may primarily optimize within the overlapped region $R_{MP}$, increasing the risk of getting trapped in local minima. Thus, we use RegBN to separate the latent encoding $z_{M}$ from $z_{P}$.  

RegBN represents one latent encoding in terms of another using a linear regression model: 
\begin{equation}\label{eq:regbn}
    z_{M}=\omega \cdot z_{P}+z_{M,r}, 
\end{equation}
in which $\omega \in{\mathbb{R}^{f\times f}}$ is a projection matrix, and $z_{M,r} \in{\mathbb{R}^{f}}$ denotes the difference between the input latent encodings, so-called residual. The residual segment contains a portion of $z_{M}$ that is independent from $z_{P}$. 
RegBN uses the Frobenius norm as a regularizer to minimize $||z_{M}-\omega \cdot z_{P}||_2$ over mini-batches. As demonstrated in ~\cref{sec:ablation_regbn}, RegBN enables a neural network to integrate data from all three non-overlapping regions $\{R_M, R_P, R_{MP}\}$, leading to improved diagnosis performance. The issue of falling into the redundant overlapping region regularly occurs in multi-modal data learning, yet is often overlooked despite its potential side effects. 
\vspace{0.5em}


In the final step, the attention maps from all three branches $\mathcal{F}_{M}$, $\mathcal{F}_{P}$, and $\mathcal{F}_{M,P}$ pass through the feed-forward network and yield their latent feature vectors respectively. 
All latent vectors of the three branches are then summed and unified through an MLP layer to obtain the diagnostic label, i.e., $\mathcal{F}:{\mathbb{R}^{f}}\rightarrow {\mathbb{R}^{L}}$. 

\subsection{Backbone Architecture}
We adopt a 3D multiple instance neuroimage Transformer~\cite{minit} as our backbone, while it is straightforward to generalize the backbone to any 3D ViT architecture. 
The adopted backbone is a convolution-free Transformer-based architecture inspired by the multiple instance learning paradigm. Given the input image, the backbone first splits it into $N_b$ non-overlapping cubiform blocks $m_i \in \mathbb{R}^{b\times b\times b}$, where $i \in \{1,...,N_b\}$, $m \in \{\mathbf{M}, \mathbf{P}\}$, and $b$ denotes the block size. Each $m_i$ is then treated as one instance to be fed into the Transformer encoder independently. Learned block positional embeddings will be added to each $m_i$, aiming to preserve the positional information of each block within the whole image. 
The Transformer encoder consists of $N$ Transformer layers. Each of the layers contains either a multi-head self-attention block or a multi-head bi-attention block, followed by a feed forward network including two linear projections with Gaussian Error Linear Units (GELU) non-linearity~\cite{hendrycks2016gaussian} applied in between. We apply layer normalization~\cite{ba2016layer} before each block, and residual connections after each block within a Transformer layer.

\section{Experiment Setup}
\label{sec:experiments}

\noindent\textbf{Dataset and Preprocessing:} We use paired FDG-PET and T1-weighted MRI scans from three datasets:
\begin{enumerate}
\vspace{-0.2em}
    \item Alzheimer’s disease neuroimaging initiative (ADNI) database~\cite{adni}, including 379 CN, 257 AD, and 611 MCI samples.
    \vspace{-0.5em}
    \item Japanese Alzheimer’s Disease Neuroimaging Initiative (J-ADNI) database~\cite{jadni}, including 104 CN, 78 AD, and 124 MCI samples.
    \vspace{-0.5em}
    \item In-house clinical dataset containing two types of dementia, with 143 CN, 110 AD, and 57 FTD samples, from Klinikum rechts der Isar, Munich, Germany.  
\end{enumerate}


\noindent 
Due to the lack of a public multi-modal dataset for both AD and FTD subjects, we evaluate the differential diagnosis on a well-characterized, single-site in-house clinical dataset from Klinikum rechts der Isar, Munich, Germany.
\cref{tab:dataset_statistics}
reports statistics of all used datasets. 
Only scans from baseline visits are selected. We extract the gray matter (GM) density maps from MRI as input. All scans were normalized, registered to the MNI152 template with $1.5 mm^3$ voxel size, and rescaled to the intensity range between 0 and 1, with the final size of 128 × 128 × 128. 

To avoid biased results due to data leakage and confounding effects, we split the data using only baseline visits and ensure that diagnosis, age, and sex are balanced across sets. Further information on data preprocessing and splitting can be found in Sec. A.1.

\begin{table}[h]
\vspace{-0.2cm}
\centering
\scriptsize
{\setlength{\tabcolsep}{0.5em}
    \caption{Statistics for our datasets.}
    \vspace{-0.2cm}
    {\begin{tabular}{lcccc}
    \toprule
        Dataset & Diagnosis & \ \# Samples & \ \% Female & Age \\
        \midrule

        \multirow{3}{*}{ADNI}
        & CN & 379 & 50.92 & 73.46 \interval{5.93}  \\   
        & AD & 257 & 40.47 & 74.41 \interval{7.89}  \\
        & MCI & 611 & 41.41 & 72.31 \interval{7.30}  \\
        \midrule

        \multirow{3}{*}{J-ADNI}
        & CN & 104 & 51.66 & 68.38 \interval{5.64}  \\   
        & AD & 78 & 57.05 & 74.03 \interval{6.53}  \\
        & MCI & 124 & 50.21 & 73.25 \interval{5.86}  \\
        \midrule
        
        \multirow{3}{*}{In-House}
        & CN & 143 & 46.85 & 64.14 \interval{9.97}  \\     
        & AD & 110 & 50.00 & 67.25 \interval{8.37}  \\
        & FTD & 57 & 38.60 & 65.25 \interval{9.66}  \\
    \bottomrule
    \end{tabular}}
    \label{tab:dataset_statistics}
    }
    \vspace{-0.5em}
\end{table}

\noindent\textbf{Baselines:} 
First, we compare \modelName\ against single-modality inputs by employing a 3D ResNet~\cite{petallyouneed} or a 3D neuroimage ViT~\cite{minit} as the backbone. 
Subsequently, we compare alternative multi-modal fusion techniques, including early, late, and middle fusion~\cite{petallyouneed}, using both backbones. In the end, we compare with the latest fusion methods Mul-T~\cite{GAO2023incomplete} and MMTFN~\cite{mmtfn2024}. \\

\noindent\textbf{Network Parameters:}
The ViT-based backbone of \modelName\ adopts a patch size of 8, a feature embedding dimension of 512, 8 attention heads, a model depth of 4, and a dropout rate of 0.0.
These parameters were selected after an exhaustive search and comparison using the validation set, with detailed information and results presented in~\cref{sec:ablation_params}. \\

\noindent\textbf{Miscellaneous:}
We implement models with PyTorch, using the AdamW~\cite{adamw} optimizer with a learning rate of $5 \times 10^{-4}$, a weight decay of $1 \times 10^{-5}$, a batch size of $16$, and cosine annealing as the learning rate scheduler. The models are trained on one NVIDIA A100 GPU with 40 GByte memory for 3,800 iterations, with early-stop to prevent overfitting. Tab.~A.2
reports the detailed hyperparameters. \\

\noindent\textbf{Evaluation:}
We perform 5-fold cross-validation, partitioning each dataset into training, validation, and test sets with ratios of 65\%, 15\%, and 20\% respectively. We train all models for three different tasks: 
\begin{enumerate}
\vspace{-0.6em}
    \item Binary classification of healthy controls (CN) vs. patients with AD.
    \vspace{-0.6em}
    \item Three-way classification of CN vs. MCI vs. AD.
    \vspace{-0.6em}
    \item Three-way classification of CN vs. AD vs. FTD.
    \vspace{-0.6em}
\end{enumerate}
We use balanced accuracy (BACC) and area under the ROC curve (AUC) to evaluate the results for binary classification. For the three-way classification, we adopt BACC, F1-Score, Precision, and Recall. In addition, we illustrate the fairness evaluation on \modelName\ across different demographics.

\section{Results}
\label{sec:results}

\begin{table*}[ht!]
\small
\vspace{-1.5em}
\caption{Results for AD prediction on two different datasets, with MRI ($\mathbf{M}$) and PET ($\mathbf{P}$) as input.}
\vspace{-1.5em}
\label{tab:cn-ad-mci-main}
\begin{center}
{\setlength{\tabcolsep}{1.2em}
\begin{tabular}{l l c cc cc }
\toprule
\multirow{2}{*}{{Data}} 
& \multirow{2}{*}{Method} & \multirow{2}{*}{Modality} & \multicolumn{2}{c}{CN vs. AD} & \multicolumn{2}{c}{CN vs. MCI vs. AD} \\\cmidrule(lr){4-7}
& & & BACC & AUC & BACC & F1-Score \\
\midrule
\parbox[t]{2mm}{\multirow{12}{*}{ADNI}}
& 3D-ResNet~\cite{petallyouneed} & $\mathbf{M}$ & 86.59 \interval{4.22} & 93.67 \interval{2.43} & 53.58 \interval{6.20} & 53.70 \interval{6.16} \\
& 3D-ResNet~\cite{petallyouneed} & $\mathbf{P}$ & 89.07 \interval{2.67} & 95.19 \interval{2.22} & 54.84 \interval{3.40} & 55.79 \interval{4.22} \\
& 3D-ViT~\cite{minit} & $\mathbf{M}$ & 86.22 \interval{3.83} & 93.66 \interval{2.50} & 60.83 \interval{7.41} & 60.03 \interval{7.45} \\
& 3D-ViT~\cite{minit} & $\mathbf{P}$ & 88.75 \interval{1.69} & 93.53 \interval{1.39} & 58.24 \interval{4.64} & 53.20 \interval{4.41} \\
& ResNet-based early-fusion\cite{earlyfusion} & $\mathbf{M}$+$\mathbf{P}$ & 82.66 \interval{2.40} & 85.58 \interval{7.29} & 57.30 \interval{2.30} & 55.79 \interval{1.02} \\
& ResNet-based middle-fusion\cite{petallyouneed} & $\mathbf{M}$+$\mathbf{P}$ & 82.63 \interval{7.04} & 88.17 \interval{6.77} & 53.01 \interval{3.40} & 53.70 \interval{6.16} \\
& ResNet-based late-fusion\cite{petallyouneed} & $\mathbf{M}$+$\mathbf{P}$ &  89.74 \interval{1.98} & 96.73 \interval{0.92} & 57.71 \interval{2.90} &  58.89 \interval{1.73} \\
& ViT-based early-fusion & $\mathbf{M}$+$\mathbf{P}$ &  89.20 \interval{3.29} & 94.87 \interval{1.80} & 62.89 \interval{2.08} & 59.84 \interval{0.73}  \\
& ViT-based late-fusion & $\mathbf{M}$+$\mathbf{P}$ &  90.60  \interval{3.57} & 96.48 \interval{1.24} & 61.86 \interval{3.00} & 59.60 \interval{5.83} \\
& Mul-T~\cite{GAO2023incomplete} & $\mathbf{M}$+$\mathbf{P}$ & 86.37 \interval{3.25} & 93.59 \interval{1.54} &  56.49 \interval{3.82} & 55.54 \interval{3.94} \\
& MMTFN~\cite{mmtfn2024} & $\mathbf{M}$+$\mathbf{P}$ &  88.76  \interval{1.98} & 93.69 \interval{1.95} &  63.11 \interval{4.51} & 60.51 \interval{3.49}
\\\cmidrule(lr){2-7}
& \textbf{\modelName~(Ours)} & $\mathbf{M}$+$\mathbf{P}$ & \textbf{92.42 \interval{2.63}} & \textbf{97.11 \interval{1.47}} & \textbf{65.18 \interval{1.57}} & \textbf{64.89 \interval{2.78}} \\
\midrule

\parbox[t]{2mm}{\multirow{12}{*}{J-ADNI}}
& 3D-ResNet~\cite{petallyouneed} & $\mathbf{M}$ & 84.66 \interval{2.92} & 91.40 \interval{3.16} & 56.89 \interval{1.64} &  54.87 \interval{1.83} \\
& 3D-ResNet~\cite{petallyouneed} & $\mathbf{P}$ & 85.48 \interval{9.87} & 92.22 \interval{4.51} & 50.87 \interval{12.8} & 46.16 \interval{18.7} \\
& 3D-ViT~\cite{minit} & $\mathbf{M}$ & 83.65 \interval{4.15} & 92.26 \interval{1.29} & 55.11 \interval{1.62} & 47.71 \interval{5.23} \\
& 3D-ViT~\cite{minit} & $\mathbf{P}$ & 89.42 \interval{3.08} & 96.14 \interval{1.16} & 51.77 \interval{5.45} & 51.32 \interval{4.74} \\
& ResNet-based early-fusion\cite{earlyfusion} & $\mathbf{M}$+$\mathbf{P}$ & 86.02 \interval{6.85} & 91.61 \interval{3.82} & 53.46 \interval{3.92} & 53.28 \interval{3.70} \\
& ResNet-based middle-fusion\cite{petallyouneed} & $\mathbf{M}$+$\mathbf{P}$ & 77.73 \interval{8.97} & 74.12 \interval{6.78} & 49.48 \interval{2.85} & 43.94 \interval{7.39} \\
& ResNet-based late-fusion\cite{petallyouneed} & $\mathbf{M}$+$\mathbf{P}$ &  88.81 \interval{3.36} & 94.85 \interval{1.79} & 50.20 \interval{6.23} & 48.90 \interval{6.50} \\
& ViT-based early-fusion & $\mathbf{M}$+$\mathbf{P}$ &  89.59 \interval{3.68} & 93.86 \interval{2.98} & 54.68 \interval{5.32} & 52.25 \interval{3.42}  \\
& ViT-based late-fusion & $\mathbf{M}$+$\mathbf{P}$ &  86.70 \interval{7.73} & 96.34 \interval{2.42} & 54.07 \interval{5.80} & 50.83 \interval{4.89} \\
& Mul-T~\cite{GAO2023incomplete} & $\mathbf{M}$+$\mathbf{P}$ & 81.02 \interval{6.08} & 88.45 \interval{2.96} & 50.04 \interval{7.28} & 45.66 \interval{8.46} \\
& MMTFN~\cite{mmtfn2024} & $\mathbf{M}$+$\mathbf{P}$ &  86.74 \interval{6.05} & 84.27 \interval{7.91} &  57.55 \interval{2.81} & 53.45 \interval{6.97}
\\\cmidrule(lr){2-7}
& \textbf{\modelName~(Ours)} & $\mathbf{M}$+$\mathbf{P}$ & \textbf{91.72 \interval{2.52}} & \textbf{96.20 \interval{2.50}} & \textbf{58.44 \interval{3.61}} & \textbf{58.88 \interval{1.73}} \\
\bottomrule
\end{tabular}}
\end{center}
\vspace{-1.8em}
\end{table*}

\subsection{Alzheimer's Prediction}
We compare \modelName\ with other baseline methods for the task of Alzheimer's prediction on two different datasets. \cref{tab:cn-ad-mci-main} reports the results for both binary (CN vs. AD) and three-way (CN vs. MCI vs. AD) classification on ADNI and J-ADNI datasets, respectively. For binary classification, PET scans generally yield better results than MRI as a single modality input, reaching a BACC of 89\% on ADNI and 89.4\% on J-ADNI, whereas MRI only achieves a BACC of 86.6\% on ADNI and 84.7\% on J-ADNI at the largest. Early, middle, and late fusion methods, together with Mul-T~\cite{GAO2023incomplete} and MMTFN~\cite{mmtfn2024}, regardless of using ResNet or ViT as the backbone structure, can only achieve on-par performance as PET alone, which is aligned with the conclusion in~\cite{petallyouneed}, as the diagnostic performance of existing multi-modal fusion methods may not yet outperform that of using PET alone. 
On the contrary, \modelName\ outperforms all other methods and single modality input by a large margin for both datasets, achieving a BACC of 92.4\% in ADNI, and 91.7\% in J-ADNI.

As for the three-way classification between CN, AD, and MCI, despite an overall performance decline due to the complexity of MCI as a syndrome, \modelName~consistently surpasses all other methods, achieving a notable BACC of 65.2\% on the ADNI dataset. 
The results highlight the effectiveness of the multi-modal fusion mechanisms employed in \modelName. The high and consistent accuracy is likely achieved through the efficient feature extraction and integration enabled by the attention mechanisms in our ViT-based framework.

\subsection{Differential Diagnosis of Dementia} 
Further, we conduct a three-way differential diagnosis of dementia, between subjects of CN, AD, and FTD. This task is challenging due to the overlapping symptoms between different types of dementia; however, it is highly important due to its distinctive clinical value~\cite{chouliaras2023use}. 
\cref{tab:ad-ftd-main} reports the results of \modelName\ and other baseline methods on this task.

When using single modalities as input, MRI only achieves a BACC of 56.7\% with ResNet, and 66.0\% using ViT as backbone. PET can achieve a BACC of 68.7\% with ResNet and 69.5\% with ViT, confirming its higher sensitivity in the differential diagnosis. Using multi-modal input of both MRI and PET can elevate the diagnostic accuracy compared to using single modality alone, particularly when employing the late fusion strategy using either ResNet or ViT, achieving a BACC of 73.2\% and 74.1\%, respectively. 
Recent fusion methods Mul-T~\cite{GAO2023incomplete} and MMTFN~\cite{mmtfn2024} combine CNNs for feature extraction with ViTs for fusion, however, their performance falls short compared to pure ViT-based late fusion.
Notably, utilizing ViT as the backbone consistently outperforms ResNet in differential diagnosis, whether employing a single modality or multiple modalities as input.
In the end, applying \modelName\ further boosts the diagnostic accuracy to more than 2\%, reaching the highest BACC of 76.5\%. These outcomes confirm the efficacy of Transformers, particularly the efficient use of attention mechanisms in \modelName, in effectively integrating multi-modal data for the challenging task of differential diagnosis of dementia, highlighting its substantial clinical value.

\begin{table*}[t]
\small
\vspace{-1.em}
\caption{Differential diagnosis between CN, AD, and FTD using MRI ($\mathbf{M}$) and PET ($\mathbf{P}$) as input on the in-house multi-dementia dataset.}
\vspace{-1.em}
\label{tab:ad-ftd-main}
\begin{center}
{\setlength{\tabcolsep}{1.2em}
\begin{tabular}{l c cc cc }
\toprule
 \multirow{2}{*}{Method} & \multirow{2}{*}{Modality} & \multicolumn{4}{c}{CN vs. AD vs. FTD} \\\cmidrule{3-6}
& & BACC & F1-Score & Precision & Recall \\
\midrule
 3D-ResNet~\cite{petallyouneed} & $\mathbf{M}$ & 56.73 \interval{4.12} & 55.32 \interval{5.28} & 64.20 \interval{5.62} & 56.62 \interval{4.92} \\
 3D-ResNet~\cite{petallyouneed} & $\mathbf{P}$ & 68.68 \interval{5.84} & 69.07 \interval{6.81} & 69.97 \interval{3.96} & 68.23 \interval{5.07}  \\
 3D-ViT~\cite{minit} & $\mathbf{M}$ & 65.99 \interval{6.58} &  62.58 \interval{7.81} & 63.46 \interval{8.50} & 65.30 \interval{5.28} \\
 3D-ViT~\cite{minit} & $\mathbf{P}$ & 69.47 \interval{5.05} &  69.53 \interval{5.84} & 69.81 \interval{3.88} & 69.20 \interval{5.87} \\
 ResNet-based early-fusion~\cite{earlyfusion} & $\mathbf{M}$+$\mathbf{P}$ & 55.06 \interval{6.55}  & 53.10 \interval{6.61} & 61.72 \interval{6.07} & 54.87 \interval{5.03} \\
 ResNet-based middle-fusion~\cite{petallyouneed} & $\mathbf{M}$+$\mathbf{P}$ & 50.83 \interval{11.3} & 43.64 \interval{15.5} & 55.12 \interval{12.6} & 51.56 \interval{11.4}  \\
 ResNet-based late Fusion~\cite{petallyouneed} & $\mathbf{M}$+$\mathbf{P}$ &  73.19 \interval{3.81} & \underline{73.87 \interval{4.26}} & 74.90 \interval{3.44} &  73.28 \interval{4.67} \\
 ViT-based early-fusion & $\mathbf{M}$+$\mathbf{P}$ &  71.04 \interval{5.00}  & 72.32 \interval{5.40} & \underline{75.50 \interval{3.92}} &  70.89 \interval{5.63} \\
 ViT-based late-fusion & $\mathbf{M}$+$\mathbf{P}$ &  \underline{74.06 \interval{3.15}}  & 73.14 \interval{6.56} & 71.62 \interval{4.07} & \underline{74.42 \interval{6.53}} \\
 Mul-T~\cite{GAO2023incomplete} & $\mathbf{M}$+$\mathbf{P}$  &  70.84 \interval{4.27} & 70.70 \interval{3.94} & 72.43 \interval{5.01} &  72.91 \interval{4.11}
 \\
 MMTFN~\cite{mmtfn2024} & $\mathbf{M}$+$\mathbf{P}$ &  72.74 \interval{3.80} & 68.94 \interval{3.91} & 71.97 \interval{3.09} & 71.41 \interval{3.14} \\
 \cmidrule(lr){1-6}
 \textbf{\modelName~(Ours)} & $\mathbf{M}$+$\mathbf{P}$ & \textbf{76.46 \interval{3.33}} & \textbf{75.53 \interval{4.38}}  & \textbf{76.76 \interval{4.88}} & \textbf{75.39 \interval{3.23}}  \\
\bottomrule
\end{tabular}}
\end{center}
\vspace{-1.1em}
\end{table*}

\subsection{Fairness Evaluation}
Ensuring fairness is a paramount consideration in the domain of medical imaging. DL models employed in medical applications must minimize biases towards specific demographic groups, such as age, gender, and diagnostic labels. In this regard, we evaluate the fairness of the diagnostic results produced by \modelName\ on the ADNI dataset for AD prediction, by examining its test accuracy across diverse patient cohorts. 
The results presented in ~\cref{tab:fairness} indicate that \modelName\ achieves minimal variance in the diagnostic accuracy across different demographic categories, suggesting a uniform and equitable performance.
\begin{table}[h!]
\small
    \centering
    \caption{Fairness evaluation of \modelName\ on the ADNI dataset. We report true positive rate (TPR) for diagnosis.}
    \vspace{-0.4em}
    {\setlength{\tabcolsep}{1.2em}
    \begin{tabular}{ccc} \toprule
         Demographics & Groups & BACC (\%) \\\midrule
         \multirow{4}{*}{Age} & $<=$ 65 & 93.04 \interval{4.53} \\
          & 65 - 70 & 91.79 \interval{8.11} \\
          & 70 - 75 & 93.20 \interval{4.21} \\
          & 75 - 80 & 90.13 \interval{2.07} \\
          & $>$ 80 & 90.44 \interval{3.93} \\\midrule
        \multirow{2}{*}{Gender} & Male & 92.52 \interval{3.40} \\
         & Female & 91.71 \interval{3.93} \\\midrule
        \multirow{2}{*}{Diagnosis} & CN & TPR = 94.46 \\
         & AD & TPR = 89.49 \\\midrule
        \multicolumn{2}{c}{Total} & 92.42 \interval{2.63} \\
       \bottomrule
    \end{tabular}}
    \label{tab:fairness}
    \vspace{-0.5em}
\end{table}

\section{Ablation Study}
We conduct a comprehensive ablation study on the important components in \modelName. This includes evaluating the inclusion of different ViT branches (self- and bi-attention), the integration of RegBN, and the application of the attention threshold $\tau$ in our bi-attention design. In the end, we include the ablation on the network parameters.

\subsection{Different Branches}
\label{sec:ablation_branches}
\modelName~comprises three independent ViT branches: $\mathcal{F}_{M}$ with MRI as input, $\mathcal{F}_{P}$ with PET, and $\mathcal{F}_{M,P}$ receives both MRI and PET as input simultaneously.
To validate the efficacy of the three branches in \modelName, we conduct ablation studies on each of the branch and their different combinations. As shown in~\cref{tab:ablation_branches}, using $\mathcal{F}_{M}$ or $\mathcal{F}_{P}$ alone achieves a BACC of 86.2\% and 88.8\%, respectively, indicating that PET is slightly more effective as a single modality input. The model achieves a BACC of 90.87\% when it relies solely on the $\mathcal{F}_{M,P}$ branch, suggesting the high benefits from our introduced bi-attention mechanism for integrating multiple modalities. The efficacy of the combination of multiple attention mechanisms is further evidenced when $\mathcal{F}_{M,P}$ is combined alongside either $\mathcal{F}_{M}$ or $\mathcal{F}_{P}$, resulting in BACC scores of 91.0\% and 91.5\%, respectively. Finally, combining all three branches, as in our final \modelName\ framework, achieves the highest BACC of 92.4\%, confirming the synergistic effect of \modelName\ with self- and bi-attention interaction.

\begin{table}[!h]
\small
\vspace{-0.4em}
    \centering
    \caption{Ablation study on the performance of different branches in \modelName. Results are on 5-fold cross-validation for binary CN vs. AD classification using ADNI dataset.}
    \vspace{-0.5em}
    {\setlength{\tabcolsep}{0.8em}
    \begin{tabular}{c c c c c c}
    \toprule
         $\mathcal{F}_{M}$  & $\mathcal{F}_{P}$ & $\mathcal{F}_{M,P}$ & Modality &BACC (\%) \\
        \midrule
        \cmark & $\times$ & $\times$ & $\mathbf{M}$ & 86.22 \interval{3.83}  \\ 
        $\times$ & \cmark & $\times$ & $\mathbf{P}$ & 88.75 \interval{1.69}  \\ 
        $\times$ & $\times$ & \cmark & $\mathbf{M}$+$\mathbf{P}$  & 90.87 \interval{3.68}  \\ 
        \cmark & \cmark & $\times$ & $\mathbf{M}$+$\mathbf{P}$ & 91.02 \interval{2.48}  \\ 
        \cmark & $\times$ & \cmark & $\mathbf{M}$+$\mathbf{P}$ & 91.77 \interval{2.05}  \\ 
        $\times$ & \cmark & \cmark & $\mathbf{M}$+$\mathbf{P}$ & 91.45 \interval{2.89}  \\ 
        \cmark & \cmark & \cmark & $\mathbf{M}$+$\mathbf{P}$ & 92.42 \interval{2.63}  \\ 
        \bottomrule
    \end{tabular}}
    
    \label{tab:ablation_branches}
    \vspace{-0.6em}
\end{table}

\subsection{Integration of RegBN}
\label{sec:ablation_regbn}
RegBN is incorporated into \modelName\ as a normalization technique to make self-attention branches independent, aiming to reduce the redundant partial dependency between the input modalities. 
We further evaluate the impact of RegBN in our model. As shown in ~\cref{tab:regbn}, the performance of \modelName\ is affected by the presence of RegBN across all three datasets, with this normalization method enhancing classification results by up to 2\%.
This improvement underscores two key issues in multi-modal classification. First, partial dependency and confounders in multi-modal data can mislead a classification neural network, causing it to fall into the overlapped region. Second, RegBN demonstrates a unique ability to counteract the negative impact of overlapping data, thereby contributing to a more efficient and robust framework. 
As discussed in previous sections, the classification of multi-modal data may differ from that of single-modal data due to the heterogeneous nature of multi-modal data sources, which can exhibit positive or negative correlations affecting the distributions of learned features~\cite{soleymani2017survey,lu2021metadata}.  
This study demonstrates and substantiates that accounting for the data space of each modality in classification leads to significant improvements in results. This topic has received limited attention and study thus far. Importantly, recent deep learning methods have shown an increased focus on multi-modal analysis, and the strategy employed in our model for managing multi-modal data demonstrates significant potential for future research.

\begin{table}[h!]
\centering
\vspace{-0.5em}
\small
\caption{Ablation on RegBN.}
\vspace{-0.6em}
{\setlength{\tabcolsep}{0.4em}
\label{tab:regbn}
\begin{tabular}{l ccc }
\toprule
 \multirow{2}{*}{BACC}& \multicolumn{2}{c}{\scriptsize{CN vs. MCI vs. AD}} & \multirow{2}{*}{\scriptsize{CN vs. AD vs. FTD} } \\\cmidrule{2-3}
 & ADNI & J-ADNI &  \\
\midrule
w/o RegBN & 63.97 \interval{3.11} & 56.35 \interval{3.49} & 74.32 \interval{5.80}  \\
w/ ~RegBN & \textbf{65.18 \interval{1.57}} & \textbf{58.44 \interval{3.61}} & \textbf{76.46 \interval{3.33}} \\
\bottomrule
\end{tabular}}
\vspace{-1.em}
\end{table}

\subsection{Bi-Attention Threshold $\tau$}
\label{sec:ablation_tau}
\begin{figure}[h]
    \centering
    \vspace{-1em}
    \includegraphics[width=0.8\linewidth]{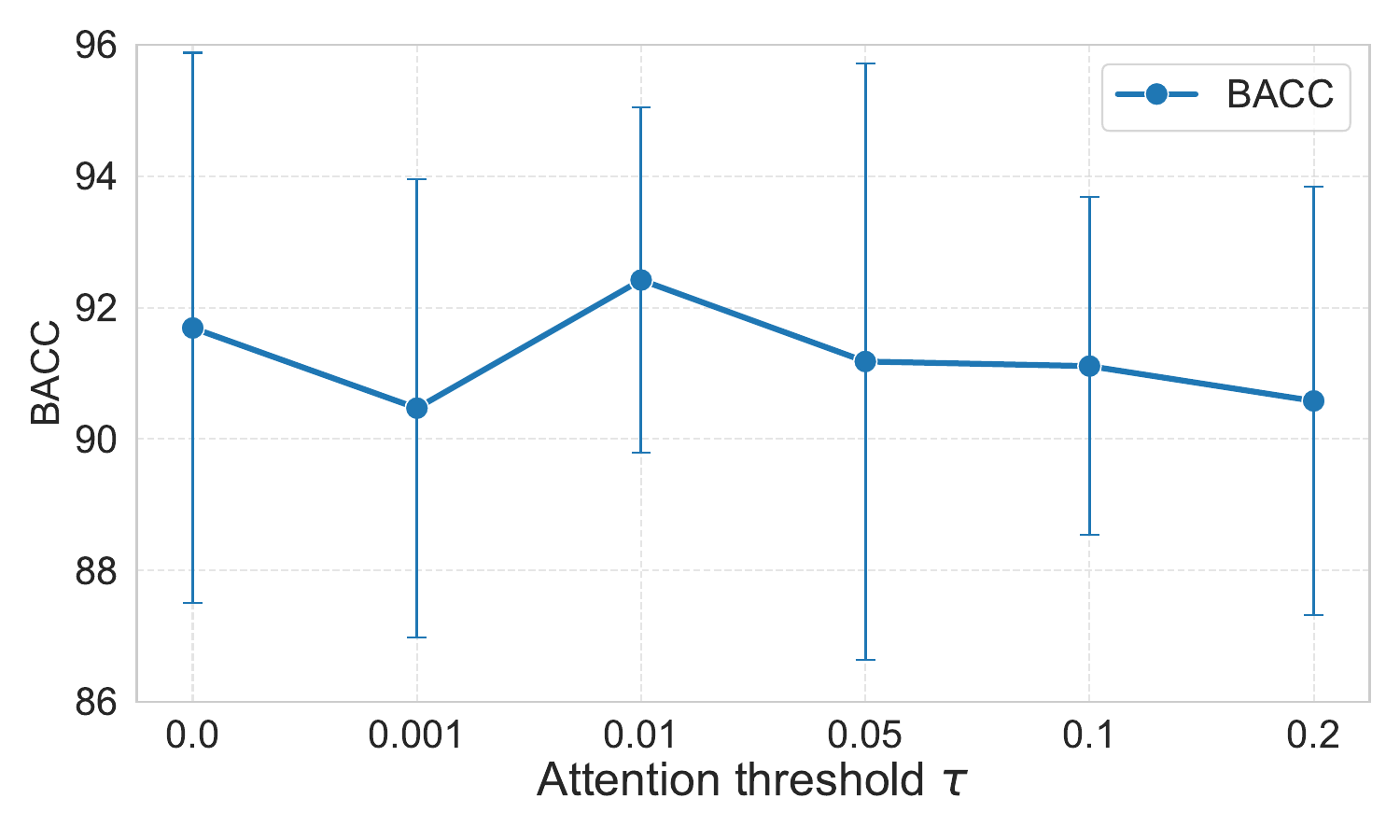}
    \caption{Ablation on the attention threshold $\tau$. Results are on the classification of CN vs. AD using the ADNI dataset.}
    \label{fig:Tau_ablation}
    \vspace{-1em}
\end{figure}

We use a constant threshold $\tau$ in~\cref{eq:biattention} to filter out very small values in the correlation matrices within the bi-attention block, so that it focuses primarily on similarities between modalities. As illustrated in~\cref{fig:Tau_ablation}, a $\tau$ value of approximately 0.01 typically results in better and more stable performance. In contrast, having no threshold ($\tau=0$, equals to a conventional cross-attention mechanism) or setting the threshold too high causes a drop in performance or high variation in the outcomes. Thus, including an optimal attention threshold is crucial for the bi-attention block, as it reduces the redundancy of learning repetitive features as captured in the self-attention blocks, and efficiently helps to focus on the dependencies between modalities.

\subsection{Network Parameters}
\label{sec:ablation_params}
We conduct ablation studies on the network parameters, performing an exhaustive search over the following parameters: patch size in \{4, 8\}, embedding dimension in \{128, 256, 512, 1024\}, number of attention heads in \{8, 16\}, model depth in \{1, 2, 4, 8\}, and dropout rate in \{0, 0.2, 0.5\}. 
We use the validation set to compare different configurations, with the results presented in ~\cref{tab:ablation_params}. As a result, the combination of a patch size of 8, a feature dimension of 512, 8 heads, a depth of 4, and a dropout rate of 0.0 yields the highest validation accuracy. Therefore, we adopt these parameters for all experiments.

\begin{table}[h]
\scriptsize
\vspace{-1.em}
\caption{Ablation on network parameters. Results are on 5-fold validation accuracy for CN vs. AD using the ADNI dataset.}
\vspace{-1.5em}
\label{tab:ablation_params}
\begin{center}
{\setlength{\tabcolsep}{0.3em}
\begin{tabular}{cc |cccc |cc |cccc |ccc | c}
\toprule
 \multicolumn{2}{c}{Patch Size} & \multicolumn{4}{c}{Embedding dim.} & \multicolumn{2}{c}{\#Heads} & \multicolumn{4}{c}{Depth}& \multicolumn{3}{c}{Dropout rate} & \multirow{2}{*}{BACC (\%)} \\
4 & 8 & 
128 & 256 & 512 & 1024 &
8 & 16  &
1 & 2 & 4 & 8 &
0 & 0.2 & 0.5 &
\\\midrule
\checkmark &  &    
\checkmark &  &  &  &  
\checkmark &   &   
 & \checkmark &  &  &  
\checkmark &  &  &  
91.71 \smallinterval{1.42}  \\ 
\checkmark &  &    
\checkmark &  &  &  &   
\checkmark &   &   
 &  & \checkmark &  &  
\checkmark &  &  &   
89.25 \smallinterval{3.41} \\
 & \checkmark &    
\checkmark &  &  &  &  
\checkmark &   &   
 &  & \checkmark &  &  
\checkmark &  &  &  
92.43 \smallinterval{2.56} \\ 
 & \checkmark &    
 & \checkmark &  &  &  
\checkmark &   &   
 &  & \checkmark &  &  
\checkmark &  &  &  
93.19 \smallinterval{2.55} \\  
 & \checkmark &    
 &  & \checkmark &  &  
\checkmark &   &   
 &  & \checkmark &  &  
\checkmark &  &  &  
\textbf{93.75 \smallinterval{2.52}}   \\  
 & \checkmark &    
 &  &  &  \checkmark &  
\checkmark &   &   
 &  & \checkmark &  &  
\checkmark &  &  &  
90.88 \smallinterval{1.81} \\
 & \checkmark &    
 &  & \checkmark &   &  
 &  \checkmark &   
 &  & \checkmark &  &  
\checkmark &  &  &  
91.79 \smallinterval{3.72}  \\ 
 & \checkmark &    
 &  & \checkmark &   &  
 \checkmark & &   
 \checkmark &  &  &  &  
\checkmark &  &  &  
89.81 \smallinterval{2.78}  \\ 
 & \checkmark &    
 &  & \checkmark &   &  
 \checkmark &  &   
  & \checkmark &  &  &  
\checkmark &  &  &  
91.61 \smallinterval{1.23}  \\ 
 & \checkmark &    
 &  & \checkmark &   &  
 \checkmark &  &   
  & &  & \checkmark  &  
\checkmark &  &  &  
91.90 \smallinterval{3.20}  \\
 & \checkmark &    
 &  & \checkmark &   &  
 \checkmark &  &   
  & & \checkmark &   &  
 & \checkmark &  &  
 91.79 \smallinterval{3.32} \\
 & \checkmark &    
 &  & \checkmark &   &  
 \checkmark &  &   
  & & \checkmark &   &  
 &  & \checkmark &  
 91.60 \smallinterval{3.54}   \\

 & \checkmark &    
 \checkmark &  &  &   &  
 \checkmark &  &   
  & \checkmark &  &   &  
 \checkmark &  &  &  
 90.72 \smallinterval{1.91}   \\
 & \checkmark &    
  & \checkmark &  &   &  
 \checkmark &  &   
  & \checkmark &  &   &  
 \checkmark &  &  &  
 92.69 \smallinterval{2.36}   \\

 & \checkmark &    
 & \checkmark &  &   &  
  &  \checkmark&   
  & & \checkmark &   &  
 \checkmark &  &  &  
 91.79 \smallinterval{3.72}   \\
& \checkmark &    
 \checkmark &  &  &   &  
  &  \checkmark&   
  & \checkmark&  &   &  
 \checkmark &  &  &  
  91.40 \smallinterval{3.21}   \\
& \checkmark &    
  & \checkmark &  &   &  
  &  \checkmark&   
  & \checkmark&  &   &  
 \checkmark &  &  &  
  91.15 \smallinterval{1.92}   \\
  & \checkmark &    
  &  &  \checkmark &   &  
  &  \checkmark&   
  & \checkmark&  &   &  
 \checkmark &  &  &  
  92.12  \smallinterval{3.40}   \\
\bottomrule
\end{tabular}}
\end{center}
\vspace{-1.5em}
\end{table}

\section{Conclusion}
We introduced \modelName, a ViT-based framework for Alzheimer's prediction and differential diagnosis of dementia using MRI and PET. \modelName\ effectively learns from multi-modal data via self- and a novel bi-attention mechanism from a pure ViT backbone. The self-attention mechanism extracts distinct features from individual modalities, along with a normalization strategy to ensure feature independence; our novel bi-attention mechanism exclusively focuses on the similarities between multiple modalities, aiming to capture their disease-specific dependency. Across three distinct datasets, \modelName\ consistently outperformed all competing methods, achieving a balanced accuracy of 92.4\% for AD-CN classification, 65.2\% for AD-MCI-CN classification, and 76.5\% for differential diagnosis between AD, FTD, and CN subjects, highlighting its significant clinical value. We evaluated the fairness of our model, which indicated equitable performance across various demographic groups. 
Our comprehensive ablation study validated \modelName's robustness and synergistic effect of integrating multiple modalities with its intricate design. 
Overall, \modelName\ demonstrated that leveraging the attention mechanisms in vision Transformers offers superior fusion compared to CNNs~\cite{petallyouneed}, enabling the combination of MRI and PET to significantly surpass the accuracy of PET alone. 


\section*{Acknowledgements}
This work was supported by the Munich Center for Machine Learning (MCML) and the German Research Foundation (DFG). 
The authors gratefully acknowledge PD Dr. Igor Yakushev and PD Dr. Dennis M. Hedderich from Klinikum rechts der Isar (Munich, Germany) for their invaluable provision of the in-house clinical data, as well as the Leibniz Supercomputing Centre for providing the computational and data resources.


{\small
\bibliographystyle{ieee_fullname}
\bibliography{egbib}
}




\end{document}